\documentclass[
]{ceurart}

\usepackage{listings}
\begin{document}

\copyrightyear{2023}
\copyrightclause{Copyright for this paper by its authors.
  Use permitted under Creative Commons License Attribution 4.0
  International (CC BY 4.0).}

\conference{NeSy 2023: 17th International Workshop on Neural-Symbolic Learning and Reasoning, 3--5 July, 2023, Sienna, Italy}

\title{Continual Reasoning: Non-monotonic Reasoning in Neurosymbolic AI using Continual Learning}

% \tnotetext[1]{You can use this document as the template for preparing your
%   publication. We recommend using the latest version of the ceurart style.}

\author[1]{Sofoklis Kyriakopoulos}[%
email=sofoklis.kyriakopoulos@city.ac.uk,
]
\cormark[1]
% \address[1]{Department of Computer Science, City, University of London, London, UK}

\author[1]{Artur S. d'Avila Garcez}[%
email=a.garcez@city.ac.uk,
]
\address[1] {Department of Computer Science, City, University of London, London, UK}

\cortext[1]{Corresponding author.}
% \fntext[1]{These authors contributed equally.}

\begin{abstract}
  Despite the extensive investment and impressive recent progress at reasoning by similarity, deep learning continues to struggle with more complex forms of reasoning such as non-monotonic and commonsense reasoning. Non-monotonicity is a property of non-classical reasoning typically seen in commonsense reasoning, whereby a reasoning system is allowed (differently from classical logic) to \emph{jump to conclusions} which may be retracted later, when new information becomes available. Neural-symbolic systems such as Logic Tensor Networks (LTN) have been shown to be effective at enabling deep neural networks to achieve reasoning capabilities. In this paper, we show that by combining a neural-symbolic system with methods from continual learning, LTN can obtain a higher level of accuracy when addressing non-monotonic reasoning tasks. Continual learning is added to LTNs by adopting a curriculum of learning from knowledge and data with recall. We call this process \emph{Continual Reasoning}, a new methodology for the application of neural-symbolic systems to reasoning tasks. Continual Reasoning is applied to a prototypical non-monotonic reasoning problem as well as other reasoning examples. Experimentation is conducted to compare and analyze the effects that different curriculum choices may have on overall learning and reasoning results. Results indicate significant improvement on the prototypical non-monotonic reasoning problem and a promising outlook for the proposed approach on statistical relational learning examples.
  
\end{abstract}


\maketitle

\section{Introduction}
   
    The combination of machine learning and symbolic reasoning, now embodied by the area known as neurosymbolic AI, has been a developing field of research since the early days of AI. Recent advancements in deep learning allowed for a surge of interest in this particular type of models. Many variations of neural-symbolic (NeSy) models have surfaced in the past few years, showing the advantages of NeSy systems at reasoning and learning with increased explainability, data efficiency and generalization in comparison with other deep learning models \cite{garcez_3rdwave,zhang_neural_2021,besold_neural_symbolic_2017,Mao2019}.

    In this paper we propose \textbf{Continual Reasoning}, a new paradigm of learning for NeSy models to achieve \textit{non-monotonic reasoning} (NMR). The core principle of Continual Reasoning states that reasoning tasks, especially those of a non-monotonic nature, should be addressed by learning from data and knowledge in a multi-stage curriculum of training. We illustrate this learning paradigm using a combination of Logic Tensor Networks (LTN) \cite{badreddine_ltn_2020}, a NeSy framework capable of simulating First-Order Logic (FOL), and methodologies borrowed from Continual Learning (CL) for deep learning \cite{mundt_wholistic_2020}. LTN is chosen for its ability to constrain the loss calculations of a deep learning system based on symbolic knowledge defined in FOL and its effectiveness in dealing with both typical deep learning and reasoning tasks \cite{badreddine_ltn_2020,Serafini2016,Donadello2017_ltn_sii}. CL, that is, the sequential learning of knowledge, without forgetting, from data that may no longer be available, will be shown to implement non-monotonicity in LTNs efficiently, when adopting an appropriate curriculum learning. Continual Reasoning combining LTN and CL aims to address the difficulties that many NeSy models have when dealing with non-monotonic tasks.
      
    We apply and evaluate Continual Reasoning on an exemplar NMR task (the birds and penguins example), on the Smokers and Friends statistical relational reasoning task \cite{Richardson2006MLN}, and on a Natural Language Understanding (NLU) task that contains NMR (from the bAbI dataset \cite{weston2015bAbI}). Results indicate that a considerable increase in accuracy can be achieved in comparison with a single-stage curriculum of learning. 
    
    The remainder of this paper is organised as follows. In Section \ref{sec:background}, we discuss the challenges faced by previous approaches to NMR. In Section \ref{sec:method}, we introduce the Continual Reasoning methodology and two general approaches to curriculum design. In Section \ref{sec:results}, we analyze the experimental results. Section \ref{sec:discussion} concludes the paper and discusses directions for future work.

\section{Background} \label{sec:background}

    A common scenario to explain NMR is the \textit{Penguin Exception Task} (PET) \cite{Garcez2009NeuralSymbolicCR, Antoniou_NMR1997}, which can be defined in simple terms as: \textit{In a group of animals, there exist birds and non-birds. It is known that normally all birds fly, and that all non-birds do not fly. However, it is also known that penguins are animals that are birds, but do not fly}. In First-Order Logic (FOL), the PET can be defined using axioms such as $\forall X (is\_bird(X) \rightarrow can\_fly(X))$ and $\exists X( is\_penguin(X) \wedge is\_bird(X))$, etc. The idea is that in the absence of further information, it is reasonable to assume that all birds can fly. Although, when faced with information about penguins as an exception to the rule, one would like to retract the previous conclusion. In monotonic FOL, however, retracting a conclusion is not possible. Thus, in classical logic, the PET becomes unsolvable due to the contradiction that may arise from $can\_fly(X)$ and $\neg can\_fly(X)$. The PET is unsolvable also in traditional \textit{logic programming languages}, such as PROLOG \cite{covington_prolog2009}. In order to address the problem, many non-monotonic approaches have been developed, including Moore's Autoepistemic logic, McCarthy's Circumscription, Reiter's Default Logic and in logic programming with negation by failure \cite{Antoniou_NMR1997}. In autoepistemic logic, certain rules can be adjusted to include an exception: $\forall_X\ (is\_bird(X) \wedge \neg\ is\_penguin(X) \rightarrow can\_fly(X))$. However, the need to be explicit in including all exceptions makes this approach computationally expensive (considering that there are other birds that do not fly, e.g. ostriches). Circumscription and logic programming with negation by failure, on the other hand, find a solution to the problem by introducing the predicate $abnormal$, to indicate an exceptional case. The above rules would be re-written as $\forall_{X}\ (is\_bird(X) \wedge \neg\ is\_abnormal(X)) \rightarrow can\_fly(X)$ along with a rule to state that penguins are abnormal birds. Other exceptions would then be added as needed without changing the original rule. Unfortunately, this approach does not adapt well to exceptions to the exceptions such as an abnormal penguins (a hypothetical super-penguin that is capable of flying).

    At present, there is a tension between the above attempts to formalizing non-monotonicity and large-scale data-driven approaches based on neural networks and natural language that are efficient but lack any formalization. In this paper we seek to investigate approaches to solving PET and other simple examples that can be formalized but that work using the same tools as the large-scale network models. Work has been conducted to formalize NMR in neural networks starting with the Connectionist Inductive Learning and Logic Programming System (CILP) \cite{CILP}, later developed into a system for statistical relational learning. More recently, the Differentiable Inductive Logic Programming ($\partial$ILP) approach \cite{Evans_dilp2017} was proposed, addressing cycles through negation. Probabilistic approaches have also been developed which can implement a form of non-monotonicity or at least avoid the problems of classical logic by assigning probabilities to beliefs expressed as Horn clauses, e.g. DeepProbLog \cite{Manhaeve_deepproblog2018}. In this paper, rather than mapping symbolic representations into neural networks and vice-versa, we are interested in the interplay between learning and reasoning as part of a curriculum. We focus on the Logic Tensor Network (LTN) \cite{badreddine_ltn_2020} because it is a highly modular NeSy framework applicable in principle to any underlying neural network model and based on the canonical, highly expressive FOL language. Additionally, the LTN has shown promise for learning in continual mode \cite{wagner_neural_symbolic_nodate}. 
    
    The LTN relies on two main ideas, the \textit{grounding} of predicates and logical axioms into vectors and \textit{Real Logic} which maps the satisfiability of the logical axioms to a real number in the interval \{0,1\} thus enabling viewing satisfiability as optimization. Given a knowledge base of FOL axioms $\mathcal{K}$, the LTN grounds every variable $X$ to a vector representation $\mathcal{G}(X) = \langle x_1...x_k\rangle \in R^{k}$, and every predicate ${P}$ to a neural network $\mathcal{G}(p) \rightarrow [0,1]$.\footnote{A note on terminology: the LTN framework treats FOL axioms in a slightly different way than logic programming. A grounding creates a direct connection with data, mapping a variable to a specific partition of the data. For this reason, we use the term \textit{rules} instead of axioms when referring to the FOL knowledge base defined in LTN. The FOL axiom $\forall_X\ is\_bird(X) \rightarrow can\_fly(X)$ is defined in LTN as the rule $\forall_{Animals}\ is\_bird(Animals) \Rightarrow can\_fly(Animals)$, where \textit{Animals} is the set of vector groundings for all animals in the data. This makes LTN a typed FOL language. If we wish to declare rules that only apply to a subset of \textit{Animals}, we can do this in LTN using e.g. $\forall_{Norm\_Birds}\ is\_bird(Norm\_Birds)$, where \textit{Norm\_Birds} consists only of the vector representations for birds, which is a subset of \textit{Animals}. This excludes other subsets of animals, e.g. \textit{Penguins} or \textit{Cows}. For the definition of the PET used in LTN, see Appendix \ref{app:PET}.} 
    The application of \textit{Real Logic} uses differentiable fuzzy logic to calculate the truth value of any LTN rule in the usual way. The satisfiability ($sat$), i.e. the aggregated truth value of the knowledge base, is then used in the loss function, with $Loss = 1 - sat$.

\section{Method} \label{sec:method}

    \textbf{Continual Reasoning} is proposed as a novel methodology, addressing reasoning tasks with a combination of NeSy models and a curriculum of training. In CL, a multi-task dataset is split along the different tasks, so that the model can be trained on each subset of data at each stage of the curriculum, with the aim to learn new tasks without forgetting old ones. In the context of NeSy models where tasks and knowledge are mostly represented at the symbolic level, we treat the aforementioned splitting of data as a division of the symbolic knowledge along a series of stages, which constitutes our curriculum of learning. In doing so, we rely on the neural networks of the NeSy models to learn new knowledge without forgetting previously learned knowledge, adjusting their beliefs about previously learned knowledge to allow for the new knowledge to be mapped to true without creating an inconsistency. Specifically, when using the LTN as our NeSy model, a knowledge base (KB) of FOL rules is separated into multiple stages for learning. For example, consider a KB consisting of facts $a(X), b(X), c(X)$, and rules $a(X) \Rightarrow d(X)$ and $b(x) \wedge c(X) \Rightarrow d(X)$. A split into three stages might be: (1) train on the facts; (2) train on $a(X) \Rightarrow d(X)$ and recall fact $a(X)$; (3) train on $b(x) \wedge c(X) \Rightarrow d(X)$. All facts and rules are assumed to be universally quantified. Our experiments will show, as one would expect, that the choice of curriculum, i.e the specific sequence in which the rules are learned and the facts are recalled, can affect the outcome. It becomes apparent that while in traditional machine learning all data is treated equally as being i.i.d. (although recent work around out-of-distribution (OOD) learning has started to question this assumption \cite{OOD}), in reasoning tasks, especially NMR, the order in which knowledge is learned matters (in addition to the data split already identified as important in OOD learning). 
    
    Thus, we focus on two core requirements for the choice of curriculum. The first relies on the approach commonly applied in CL where data is split into separate tasks \cite{mundt_wholistic_2020}. This can be applied in Continual Reasoning by treating each predicate as an individual task and training any rule aimed at learning about said predicate in a single stage of the curriculum. We call this \textit{Task Separation}. In our previous example, we would split the KB into four stages: (1) learn $a(X)$; (2) $b(X)$; (3) $c(X)$; and (4) learn about $d(\_)$, training on both rules. 
    The second requirement takes inspiration from work conducted with knowledge graphs and lifelong learning projects such as NELL \cite{NELL-aaai15}, in which we aim to ``build up" from atomic knowledge (i.e. facts) and augment knowledge by abiding to new rules. 
    In Continual Reasoning, we can accomplish this by giving priority to learning propositional rules, and rules that are directly tied to labelled data. Following this, we aim to use rules that extend the learned domain beyond what is available to more abstract concepts. This is known as \textit{Knowledge Completion}. 
    Using again our previous example, to satisfy both requirements we would split the KB into two stages: (1) train $a(X)$, $b(X)$ and $c(X)$; (2) learn $a(X) \Rightarrow d(X)$ and $b(x) \wedge c(X) \Rightarrow d(X)$.
    
    To be able to do the above using neural networks, we must address the core issue found in CL, often referred to as \textit{catastrophic forgetting}, i.e. when the process of gradient descent leads the neural network to forget previously learned data by conforming entirely to newly provided data. To address this problem, we apply a common CL technique of \textit{rehearsal} \cite{mundt_wholistic_2020}. Rehearsal is the process by which previously seen data is sampled and recalled in the current stage of learning. For Continual Reasoning, since our knowledge is represented in FOL, in each stage of learning, we recall a random set of previously learned knowledge, such as $a(X)$ earlier, to be learned along with the current knowledge.
    
    For our analysis, we compare the task separation and knowledge completion curricula to a \textit{Baseline}, where all knowledge is learned in a single stage, and a \textit{Random} curriculum, where the KB split is randomly selected for each stage. To allow for effective comparison, all curricula, apart from the baseline, are composed of three stages. These comparisons are applied to the PET as a prototypical NMR task to show their benefits. In addition, to show the effectiveness of Continual Reasoning on other types of reasoning problems, we apply it to the Smokers and Friends task%, a statistical relational reasoning task used to benchmark reasoning capabilities of neural-symbolic models 
    \cite{badreddine_ltn_2020, riegel_logical_2020, Richardson2006MLN} and to Task 1 of the bAbI dataset \cite{weston2015bAbI} in what follows. %, as a natural language understanding task with non-monotonic reasoning. 
    
\section{Results} \label{sec:results}
    
    \begin{table*}
        \caption{Accuracy for each Curriculum Choice for the Penguin Exception Task (PET). Baseline: all rules are learned at once. Random: random split of rules along 3 stages. Task Separation (TS): divide rules according to task. Knowledge Completion (KC): divide rules to train facts before general rules. Best performing overall performance can be found in the task separation curriculum.}
        \label{tab:PET_acc_curr}
        \begin{center}
        \begin{small}

        \begin{tabular}{lccccr}
            \toprule
            Curriculum &  Rules & Stage 1 & Stage 2 & Stage 3 \\
            \midrule
            Baseline & \begin{tabular}{l}
                    is\_bird(Normal\_Birds) \\
                    is\_bird(Penguins) \\
                    can\_fly(Birds) \\
                    not(can\_fly(Penguins))
                \end{tabular} & - & - &  \begin{tabular}{c}
                    $97.1\% \pm 0.11\%$ \\
                    $61.8\% \pm 0.00\%$ \\
                    $\mathbf{96.2\% \pm 0.33\%}$ \\
                    $62.8\% \pm 0.00\%$
                \end{tabular} \\
            \midrule
            Random & \begin{tabular}{l}
                    is\_bird(Normal\_Birds) \\
                    is\_bird(Penguins) \\
                    can\_fly(Birds) \\
                    not(can\_fly(Penguins))
                \end{tabular} &
                \begin{tabular}{c}
                    $61.8\% \pm 47.8\%$ \\
                    $54.5\% \pm 45.9\%$ \\
                    $28.5\% \pm 43.9\%$ \\
                    $71.2\% \pm 43.7\%$
                \end{tabular}  &
                \begin{tabular}{c}
                    $88.2\% \pm 33.0\%$ \\
                    $58.2\% \pm 48.3\%$ \\
                    $65.7\% \pm 49.1\%$ \\
                    $44.0\% \pm 48.4\%$
                \end{tabular} &
                \begin{tabular}{c}
                    $97.7\% \pm 0.02\%$ \\
                    $67.1\% \pm 21.7\%$ \\
                    $90.5\% \pm 7.06\%$ \\
                    $79.7\% \pm 16.8\%$
                \end{tabular}  \\
            \midrule
            KC & \begin{tabular}{l}
                    is\_bird(Normal\_Birds) \\
                    is\_bird(Penguins) \\
                    can\_fly(Birds) \\
                    not(can\_fly(Penguins))
                \end{tabular} & 
                \begin{tabular}{c}
                    $99.9\% \pm 0.01\%$ \\
                    $22.5\% \pm 24.3\%$ \\
                    $57.6\% \pm 6.21\%$ \\
                    $41.4\% \pm 4.81\%$
                \end{tabular} & 
                \begin{tabular}{c}
                    $99.9\% \pm 0.01\%$ \\
                    $98.9\% \pm 2.22\%$ \\
                    $99.1\% \pm 1.94\%$ \\
                    $2.64\% \pm 5.74\%$
                \end{tabular} & 
                \begin{tabular}{c}
                    $99.9\% \pm 0.01\%$ \\
                    $99.1\% \pm 1.16\%$ \\
                    $91.9\% \pm 7.29\%$ \\
                    $78.7\% \pm 43.5\%$
                \end{tabular} \\
            \midrule
            TS & \begin{tabular}{l}
                    is\_bird(Normal\_Birds) \\
                    is\_bird(Penguins) \\
                    can\_fly(Birds) \\
                    not(can\_fly(Penguins))
                \end{tabular} & 
                \begin{tabular}{c}
                    $99.9\% \pm 0.00\%$ \\
                    $99.8\% \pm 0.02\%$ \\
                    $53.9\% \pm 5.53\%$ \\
                    $53.1\% \pm 5.25\%$
                \end{tabular} & 
                \begin{tabular}{c}
                    $99.9\% \pm 0.00\%$ \\
                    $99.9\% \pm 0.01\%$ \\
                    $99.9\% \pm 0.00\%$ \\
                    $0.01\% \pm 0.01\%$
                \end{tabular} & 
                \begin{tabular}{c}
                    $\mathbf{99.9\% \pm 0.01\%}$ \\
                    $\mathbf{99.5\% \pm 0.32\%}$ \\
                    $84.7\% \pm 2.21\%$ \\
                    $\mathbf{99.7\% \pm 0.25\%}$
                \end{tabular} \\
            \bottomrule
        \end{tabular}
        \end{small}
        \end{center}
    \end{table*}

    \begin{figure}
        \centering
        \includegraphics[width=0.9\linewidth,height=5cm]{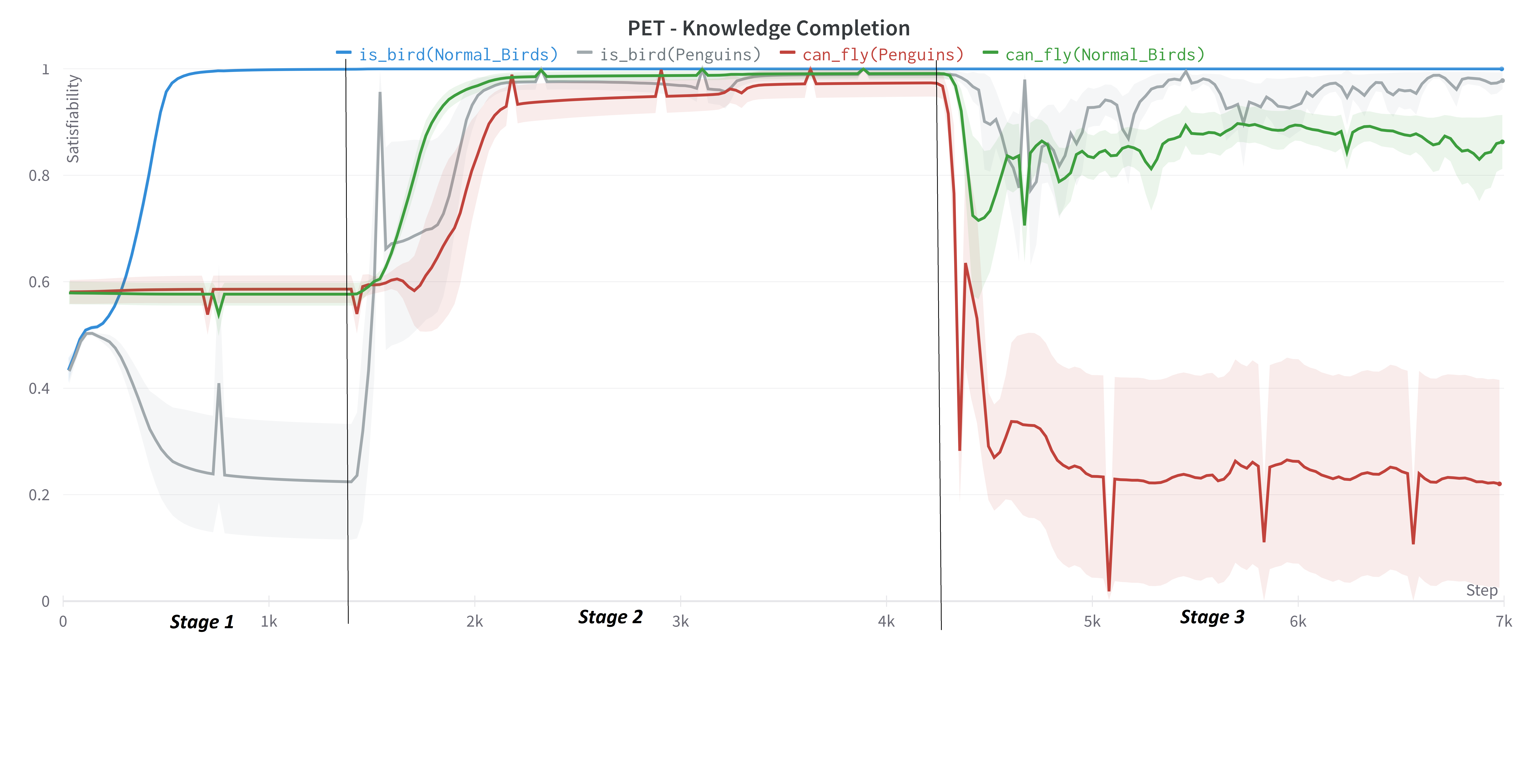}\\
        \vspace{-0.2in}
        \includegraphics[width=0.9\linewidth,height=5cm]{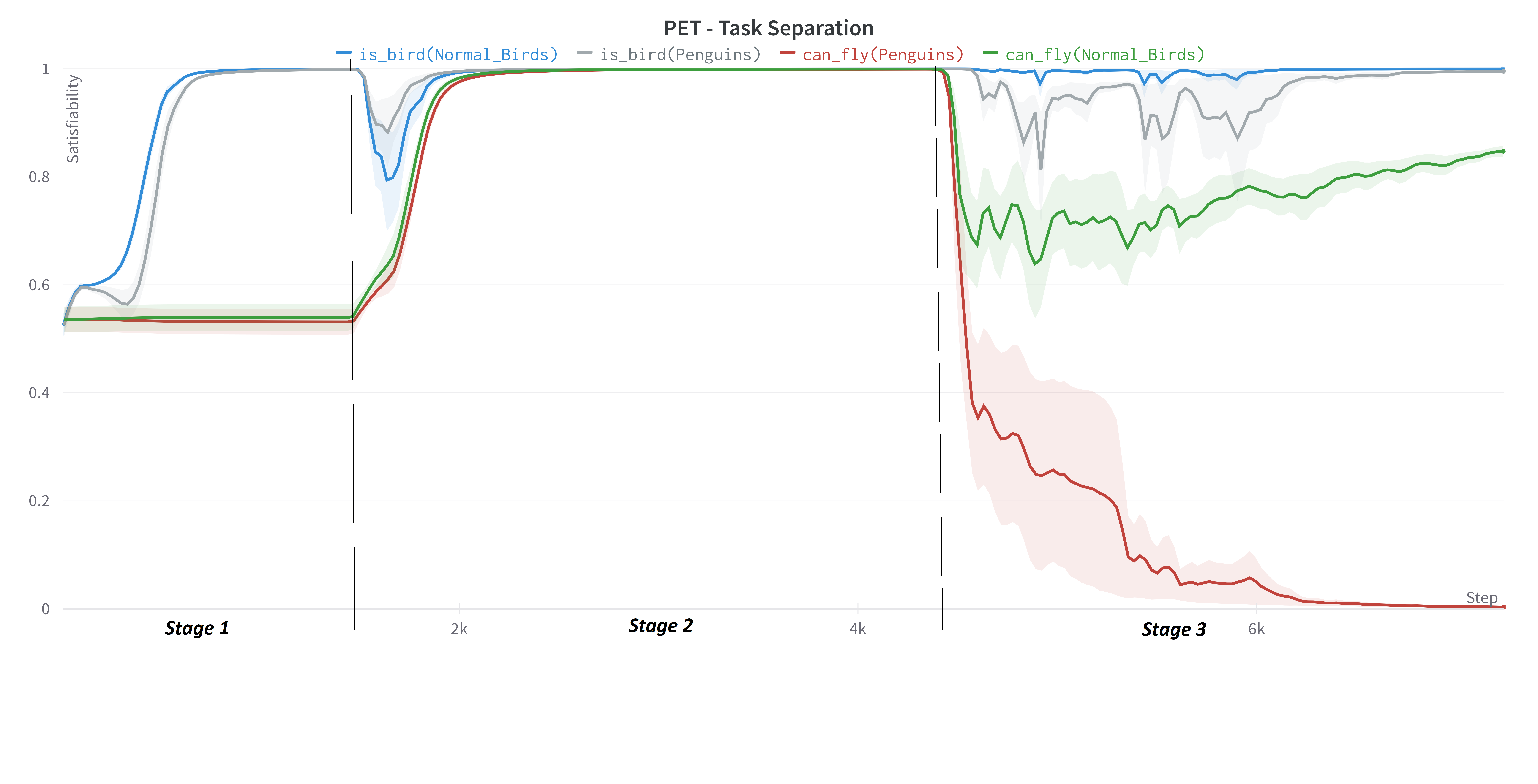} 
        \caption{Satisfiability of four LTN queries: (1) Normal birds are birds (blue), (2) Penguins are birds (grey), (3) Normal birds can fly (green), (4) Penguins can fly (red), for the Knowledge Completion curriculum (top) and the Task Separation curriculum (bottom). Overall better performance is seen in the Task Separation curriculum.}
        \label{fig:PET_Stages}
    \end{figure}

    \paragraph{Penguin Exception Task (PET)}: For the PET, we examine the behaviour of the LTN model throughout the curriculum of training, paying particular attention to three distinct types of reasoning that are necessary for success. First, we have knowledge that can be learned through induction with \textit{one-hop reasoning}, such as determining that all normal birds fly: $\forall_{Norm\_Birds}\ can\_fly(Norm\_Birds)$, and that all penguins are birds: $\forall_{Penguins}\ is\_bird(Penguins)$. Second, we have \textit{two-hop reasoning} when determining that all penguins should be able to fly, $\forall_{Penguins}\ can\_fly(Penguins)$, because they are birds. This is an instance of \textit{jumping to a conclusion} in the absence of further information.
    Lastly, we contradict this conclusion with our final learning stage for which we expect to conclude non-monotonically that penguins in fact do not fly, $\forall_{Penguins}\ \neg\ can\_fly(Penguins)$. We use these four FOL statements as queries in the analysis of our curricula of learning by measuring their LTN satifiability over time (Table \ref{tab:PET_acc_curr}).

    The results indicate that the task separation curriculum performs better than the other curricula, with the LTN able to correctly distinguish between all types of animals, as well as learn that normal birds can fly, while penguins, although still classified as birds, do not fly. The knowledge completion curriculum also performs to a high satisfiability for each of the queries. However, in comparison with task separation, the knowledge completion curriculum is less robust, and in our experimentation led to one failure case, in which penguins were misclassified as normal birds, and therefore could fly.

    When analyzing the queries throughout the training stages, we can identify changes that show that the LTN has the desired behaviour, including \textit{jumping to conclusions} and \textit{belief revision}. Specifically, in the second stage of both curricula, the LTN is trained to infer that penguins are birds, as well as that all birds can fly. Until told otherwise, the LTN jumps to the conclusion that penguins should be able to fly. In the third stage, however, the LTN is trained on the rule that penguins cannot fly. Given this knowledge, $can\_fly(Penguins)$ and $can\_fly(Normal\_Birds)$ take an initial plunge (clearly shown in Figure \ref{fig:PET_Stages}). This, of course, makes sense, as the LTN does not yet have any reason to distinguish between penguins and normal birds, and thus once again jumps to the conclusion that since penguins cannot fly, then normal birds should not fly either. However, we see that the process of recall makes $can\_fly(Normal\_Birds)$ regain satisfiability, while the satisfiability of $can\_fly(Penguins)$ decreases towards zero. It is interesting to note that in stage 3 the apparent contradiction does not lead to a convergence around an uninformative satisfiability of 0.5.  

    With a random curriculum, we see more variance in the final results, which is to be expected given the random choice of rules, but overall, on average, this curriculum performs slightly better than the baseline. This shows that even \emph{without} the benefit of curriculum design, the \emph{method of Continual Reasoning leads to better results than attempting to learn the full knowledge base in a single stage}. By further analysing the experiments in which the random curricula perform optimally, we see that task separation and knowledge completion curricula are not the only viable option for success (see Appendix \ref{app:PET}).

    \begin{table*}
        \label{tab:SF-result_comparison}
        \caption{Accuracy comparison for the Smokers and Friends statistical relational reasoning task. Compares the lower and upper bounds of LNN \cite{riegel_logical_2020} trained on $P_2^5$, and log-probability weights of MLN \cite{Richardson2006MLN} (provided in \cite{riegel_logical_2020}), for selected rules, with baseline LTN and LTN with continual learning on task separation (TS) and knowledge completion (KC) curricula.}
        \begin{center}
        \begin{small}
        
        \begin{tabular}{lccccccr}
            \toprule
            Rules & LNN-$P_2^5$ & MLN & LTN-Baseline & LTN-TS Stage 3 & LTN-KC Stage 3 \\
            \midrule
            $\neg F(x,x)$ & [0.83,0.98] & 0.26 & 0.998 & 0.996 & 0.995 \\
            $F(x,y)\Rightarrow F(y,x)$ & [0.97,1.00] & - & 0.796 & 0.826 & 0.954 \\
            $\exists_y F(x,y)$ & [1.00,1.00] & 6.88 & 0.730 & 0.748 & 0.718 \\
            $F(x,y)\wedge S(x)\Rightarrow S(y)$ & [0.65,1.00] & 3.53 & 0.716 & 0.615 & 0.497 \\
            $S(x)\Rightarrow C(x)$ & [0.58,1.00]& -1.35 & 0.715 & 0.806 & 0.978 \\
            $F(x,y)$ & - & - & 0.865 & 0.821 & 0.876 \\
            $S(x)$ & - & - & 0.825 & 0.731 & 0.653 \\
            $C(x)$ & - & - & 0.919 & 0.969 & 0.999 \\
            $\neg S(x)\Rightarrow\neg C(x)$ & - & - & 0.917 & 0.910 & 0.991 \\
            \bottomrule
        \end{tabular}
        \end{small}
        \end{center}
    
    \end{table*}

    \paragraph{Smokers and Friends Task (S\&F)}: The S\&F problem consists of a statistical relational reasoning task. We define the knowledge base in accordance with \cite{badreddine_ltn_2020} and compare a baseline curriculum to curricula belonging to knowledge completion and task separation paradigms. The satisfiability of each rule throughout the stages show that a knowledge completion curriculum outperforms the baseline and task separation on identifying that smoking causes cancer (97.8\% to 71.5\% and 80.6\%, respectively). Overall, the knowledge completion curriculum leads to the LTN reaching higher satisfiability in five of the nine FOL rules, in comparison with the baseline which beats the other curriculum in only three of the nine rules (see Appendix \ref{app:SmokerFriends} for a table detailing the satisfiability of rules per stage of each curriculum).

    In addition to comparison between curricula, we compare the outcome of Continual Reasoning with two other NeSy models that have been applied to S\&F, the Logical Neural Network (LNN) and the Markov Logic Network (MLN). The LNN allows for a lower and upper bound truth value, which signifies the lowest possible and highest possible truth value for a given FOL axiom, such that the whole knowledge base holds true. The MLN derives axiom log-probability weights which signify the probability of the axiom's mapping to true compared to the probability of it mapping to false. In Table \ref{tab:SF-result_comparison}, we see the results of these models per FOL rule used for training in our experiments. It is important to note that a precise comparison is not possible, as each model defines the set of FOL rules slightly differently in training. However, we see that the application of continual reasoning on LTNs for the S\&F task performs comparably to other NeSy approaches.

    \paragraph{bAbI - Task 1}: Task 1 of the bAbI dataset contains story lines of given facts and questions about those facts. For example, one instance will provide the sentences "Mary went to the office. Jack travelled to the garden." and ask "Where is Mary?". In order to address such a task with the proposed approach of Continual Reasoning using LTNs, we transform natural language sentences into FOL rules using GPT-3 \cite{brown_gpt2020} \footnote{This approach is inspired by that used in \cite{nye_dualsystem2021}, although FOL parsing of natural language is an evolving field of research which continues to face challenges \cite{singh_folparsing2020}}. As the task already consists of stories told in stages, separated by questions, for curriculum design, we simply separate the FOL rules along the same stages in the dataset. The reasoning here can be said to be non-monotonic over time in that, later in the story, truth-values may change, e.g. Mary may no longer be in the office. Initial experimentation showed that by applying Continual Reasoning, a LTN model achieves 96.9\% accuracy on the testing set of bAbI-Task 1, surpassing the 95\% threshold for success. Further experimentation is ongoing.

\section{Discussion, Conclusions and Future Work} \label{sec:discussion}

    We have introduced a novel methodology that integrates neurosymbolic AI and continual learning techniques in order to achieve non-monotonic reasoning. We call this Continual Reasoning, and we showed that by using Logic Tensor Networks \cite{badreddine_ltn_2020} as our neural-symbolic framework, and training the knowledge base of First-Order Logic rules in a curriculum of multiple stages, we can improve on the traditional approach of learning all rules together. Additionally, we have analysed multiple types of curricula, proposing two general paradigms for curriculum design, and showed that while even random curriculum performs better on average than the baseline, a specific design choice can allow the model to appropriately jump to conclusions and revise its beliefs more effectively.

    Experimentation conducted for this paper showed that Continual Reasoning also performs comparably on statistical relational reasoning tasks to a baseline curriculum, and other NeSy models. Continuation of this work could apply Continual Reasoning on larger datasets, such as the dataset used in RuleTaker \cite{clark_ruletaker2020}, visual relational question-answering datasets, such as the CLEVR \cite{johnson_clevr2016}, and the remaining tasks in the bAbI dataset \cite{weston2015bAbI}. 
    
    Furthermore, there still remain open questions concerning Continual Reasoning, such as how it might perform in extended non-monotonic reasoning tasks that occur when addressing lifelong learning. Rudimentary exploration of extending the PET to learn about a "super-penguin" which could fly, resulted in the LTN mostly failing to learn the exception to the exception. We believe, however, that utilising more advanced continual learning techniques, such as structural choices for neural network architecture, as well as more sophisticated recall methods like active learning, as suggested in \cite{mundt_wholistic_2020,wagner_neural_symbolic_nodate}, would allow the Continual Reasoning methodology to succeed. This is to be investigated. Additionally, while LTNs proved to be a straightforward NeSy model to apply Continual Reasoning on, it should be possible to apply our methodology to other NeSy models, such as LNNs. Integration with a very recent software framework called PyReason \cite{Aditya_pyreason2023} could provide an efficient way to do this.

% \begin{acknowledgments}
%   % Thanks to the developers of ACM consolidated LaTeX styles
%   % \url{https://github.com/borisveytsman/acmart} and to the developers
%   % of Elsevier updated \LaTeX{} templates
%   % \url{https://www.ctan.org/tex-archive/macros/latex/contrib/els-cas-templates}.  
% \end{acknowledgments}

%%
%% Define the bibliography file to be used
\bibliography{CR-references}

\newpage
\appendix

\section{Penguin Exception Task - Extra Material}
\label{app:PET}

% \subsection{PET - LTN Rule and curriculum definition}
    \paragraph{PET - LTN Rule and curriculum definition} Let us assume the variables $Norm\_Birds$, $Cows$,  $Penguins$, and $Animals$, which represent groups of normal birds, cows, penguins, and the union of all groups of animals, respectively. Therefore, we define below a knowledgebase of FOL rules that reflect the PET as a prototypical non-monotonic reasoning task.
    
    %AG: suggest you use Bird(X) notation rather than Bird(birds). Or maybe Bird(birds(X)) to denote that X is a "Bird", "birds" being a function denoting the elements of type "birds". Similarly for ¬Bird(cows), you could use Cow(X) imply ¬Bird(X), or you could use function cows(X) to denote the elements of type "cows"... what difference does it make to the LTN results to use functions or only predicates?
    \begin{enumerate}
        \item $\forall_{Norm\_Birds}\ is\_bird(Norm\_Birds)$ \hfill (normal birds are birds)
        \item $\forall_{Cows}\ \neg\ is\_bird(Cows)$ \hfill (cows are not birds)
        \item $\forall_{Animals}\ is\_bird(Animals) \Rightarrow can\_fly(Animals)$ \hfill (birds can fly)
        \item $\forall_{Animals}\ \neg\ is\_bird(Animals) \Rightarrow \neg\ can\_fly(Animals)$ \hfill (non-birds cannot fly)
        \item $\forall_{Penguins}\ is\_penguin(Penguins)$ \hfill (penguins are penguins)
        \item $\forall_{Non\_Penguins}\ \neg\ is\_penguin(Non\_Penguins)$ \hfill (non-penguins are not penguins)
        \item $\forall_{Animals}\ is\_penguin(Animals) \Rightarrow is\_bird(Animals)$ \hfill (penguins are birds)
        \item $\forall_{Animals}\ is\_penguin(Animals) \Rightarrow \neg\ can\_fly(Animals)$ \hfill (penguins do not fly)
    \end{enumerate}

    It is important to note that these rules are defined taking a open-world assumption, hence the need for declaring negations in rules 2 and 6. Additionally, we recognize that the same knowledge task could be defined using other forms of the same rules, to the same end. For example, rules 7 and 8 could be combined into one: $\forall_{Animals} is\_penguin(Animals) \Rightarrow is\_bird(Animals) \wedge \neg\ can\_fly(Animals)$. However, for the purposes of this paper, we limit the rules to their simplest forms.

    \begin{table*}[h]
        \caption{Rules selection for each stage of the curriculum for the Penguin Exception Task (PET). In the baseline curriculum, all logical rules are learning in a single stage, and in the random curriculum, the full set of rules are randomly split into three stages. Task separation (TS) and Knowledge Completion (KC) are structured according to learning paradigms. See section \ref{sec:method} for a description of each paradigm.}
        \label{tab:PET_curriculum_choice}
        \begin{center}
        \footnotesize

        \begin{tabular}{lccccr}
            \toprule
            Curriculum & Stage 1 & Stage 2 & Stage 3 \\
            % \midrule
            % \midrule
            % PET-Baseline & - & - & 
            %     \begin{tabular}{lc}
            %          normal birds are Birds\\
            %          cows are not Birds\\
            %          birds can Fly\\
            %          non-birds cannot Fly\\
            %          penguins are Penguins\\
            %          non-penguins aren't Penguins\\
            %          penguins are Birds\\
            %          penguins cannot Fly
            %     \end{tabular}\\
            % \midrule
            % PET-Random & ** & ** & ** \\
            \midrule
            PET-KC & 
                \begin{tabular}{lc}
                     normal birds are Birds\\
                     cows are not Birds\\
                     penguins are Penguins\\
                     non-penguins aren't Penguins
                \end{tabular} & 
                \begin{tabular}{lc}
                     birds can Fly\\
                     non-birds cannot Fly\\
                     penguins are Birds
                \end{tabular} & 
                \begin{tabular}{lc}
                     penguins cannot Fly
                \end{tabular}\\
            \midrule
            PET-TS & 
               \begin{tabular}{lc}
                     normal birds are Birds\\
                     cows are not Birds\\
                     penguins are Penguins\\
                     non-penguins aren't Penguins\\
                     penguins are Birds
                \end{tabular} &
                \begin{tabular}{lc}
                     birds can Fly\\
                     non-birds cannot Fly
                \end{tabular} &
                \begin{tabular}{lc}
                     penguins cannot Fly
                \end{tabular}\\
            \bottomrule
        \end{tabular}
        \normalsize
        \end{center}
    \end{table*}

% \subsection{PET - Further Results}

    \begin{figure}[h]
        \centering
        \includegraphics[width=\linewidth, height=5cm]{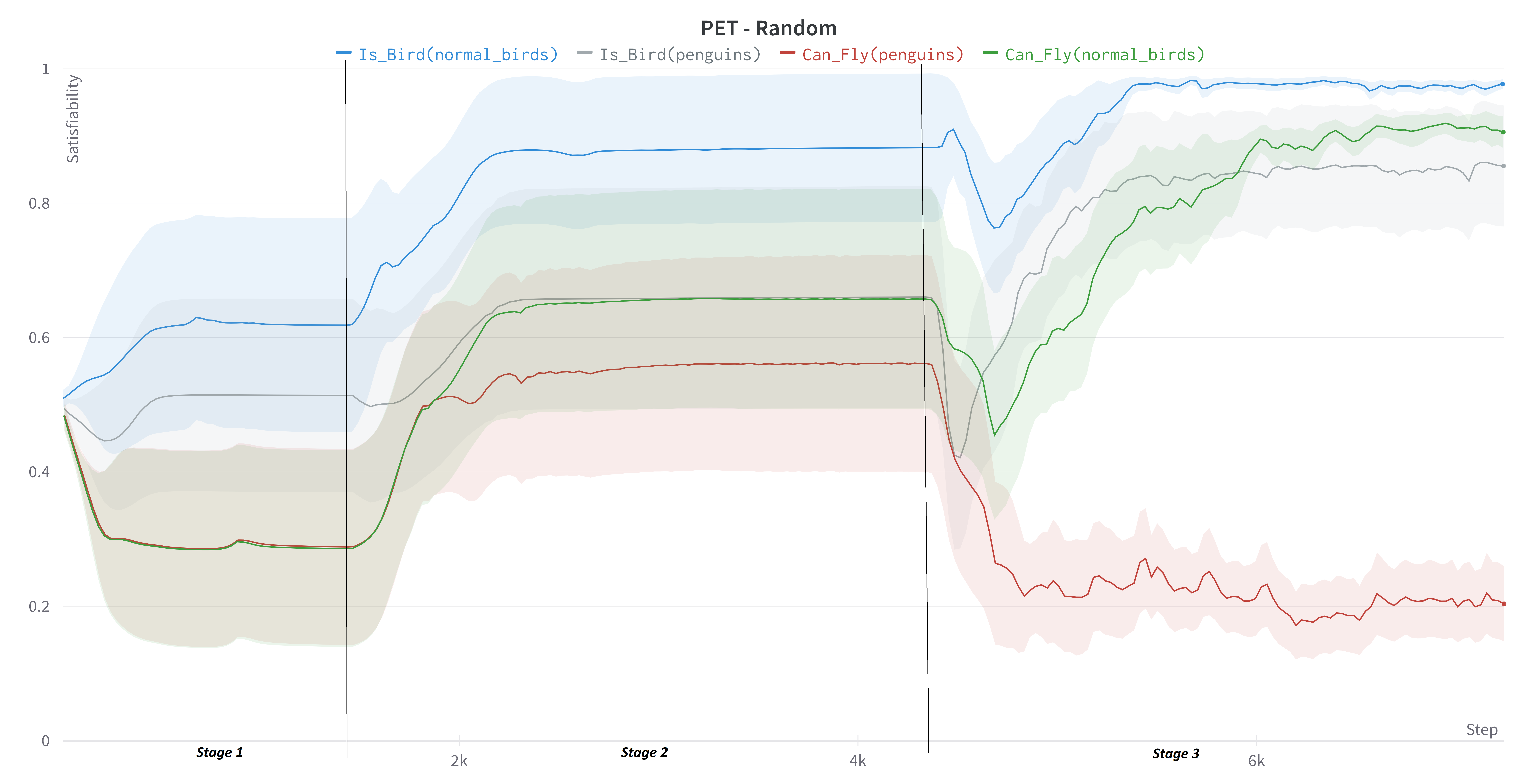}
        \caption{The figures present the satisfiability of four metrics, 1. normal birds are classified as birds (blue), 2. penguins are classified as birds (grey), 3. normal birds can fly (green), and 4. penguins can fly (red), for the Random curriculum in the PET.}
        \label{fig:PET_Stages_random}
    \end{figure}

    \begin{table}[h]
        \label{tab:PET_Random_curr_choice}
        \caption{Random Curriculum choice that leads to well performing LTN in the PET. This curriculum results in is\_bird(Normal\_Birds)=99.9\%, is\_bird(Penguins)=99.8\%, can\_fly(Normal\_Birds)=90.5\%, and not(can\_fly(Penguins))=99.8\%. One likely reason this curriculum succeeds is that the non-monotonic rule, that penguins do not fly, is learned in the final stage.}
        \begin{tabular}{lcc}
        \toprule
            Stage & FOL rules \\
            \midrule
            Stage 1 & \begin{tabular}{lc}
                 $\forall_{Animals}\ is\_bird(Animals) \Rightarrow can\_fly(Animals)$ \hfill (birds can fly) \\
                 $\forall_{Animals}\ \neg\ is\_bird(Animals) \Rightarrow \neg\ can\_fly(Animals)$ \hfill (non-birds cannot fly) \\
                 $\forall_{Non\_Penguins}\ \neg\ is\_penguin(Non\_Penguins)$ \hfill (non-penguins are not penguins) \\
                 $\forall_{Cows}\ \neg\ is\_bird(Cows)$ \hfill (cows are not birds)\\
            \end{tabular} \\
            \midrule
            Stage 2 & \begin{tabular}{lc}
                 $\forall_{Penguins}\ is\_penguin(Penguins)$ \hfill (penguins are penguins) \\
                 $\forall_{Norm\_Birds}\ is\_bird(Norm\_Birds)$ \hfill (normal birds are birds)\\
            \end{tabular} \\
            \midrule
            Stage 3 & \begin{tabular}{lc}
                 $\forall_{Animals}\ is\_penguin(Animals) \Rightarrow \neg\ can\_fly(Animals)$ \hfill (penguins do not fly) \\
                 $\forall_{Animals}\ is\_penguin(Animals) \Rightarrow is\_bird(Animals)$ \hfill (penguins are birds) \\
            \end{tabular} \\
        \bottomrule
        \end{tabular}
    \end{table}

\newpage
\section{Smokers and Friends Task - Extra Material} \label{app:SmokerFriends}

    \begin{table}[h]
        \caption{Rules selection for each stage of the curriculum for the Smokers and Friends task (S\&F). In the baseline curriculum, all logical rules are learning in a single stage. Task separation (TS) and Knowledge Completion (KC) are structured according to learning paradigms.}
        \label{tab:SF_curriculum_choice}
        \begin{center}
        \footnotesize

        \begin{tabular}{lccccr}
            \toprule
            Curriculum & Stage 1 & Stage 2 & Stage 3 \\
            \midrule
            S\&F-Baseline & - & - &
                \begin{tabular}{lc}
                     identify known friendships \\
                     identify known smokers \\
                     identify known cancer \\
                     friendship is antireflexive\\
                     friendship is symmetric \\
                     everyone has a friend \\
                     friends of smokers smoke \\
                     smokers have cancer \\
                     non-smokers don't have cancer
                \end{tabular}\\
            % \midrule
            % SF-Random & ** & ** & ** \\
            \midrule
            S\&F-KC &
                \begin{tabular}{lc}
                     identify known friendships \\
                     identify known smokers \\
                     identify known cancer
                \end{tabular} &
                \begin{tabular}{lc}
                     friendship is antireflexive\\
                     friendship is symmetric \\
                     everyone has a friend \\
                     friends of smokers smoke
                \end{tabular} &
                \begin{tabular}{lc}
                     smokers have cancer \\
                     non-smokers don't have cancer
                \end{tabular}\\
            \midrule
            S\&F-TS &
                \begin{tabular}{lc}
                     identify known friendships \\
                     friendship is antireflexive\\
                     friendship is symmetric \\
                     everyone has a friend \\
                \end{tabular} &
                \begin{tabular}{lc}
                     identify known smokers \\
                     friends of smokers smoke
                \end{tabular} &
                \begin{tabular}{lc}
                     identify known cancer \\
                     smokers have cancer \\
                     non-smokers don't have cancer
                \end{tabular}\\
            \bottomrule
        \end{tabular}
        \normalsize
        \end{center}
    \end{table}

    \begin{table}[h]
        \caption{Accuracy for each Curriculum Choice for the Smokers and Friends (S\&F) task. Baseline: all rules are learned at once. Random: random split of rules along 3 stages. Task Separation (TS): divide rules according to "task". Knowledge Completion (KC): divide rules to train facts before general rules. Asterisks signify that the associated FOL rule was trained in the particular stage of the curriculum. Best performing overall performance can be found in the knowledge completion curriculum.}
        \begin{center}
        
        \begin{tabular}{lc|ccc|cccr}
            \toprule
            Rules & Baseline & \multicolumn{3}{c}{Task Separation} & \multicolumn{3}{c}{Knowledge Completion} \\
             & & Stage 1 & Stage 2 & Stage 3 & Stage 1 & Stage 2 & Stage 3 \\
            \midrule
            $F(x,y)$ & 0.865 & \textit{0.836*} & 0.795 & 0.821 & \textit{0.872*} & 0.831 & \textbf{0.876} \\
            $S(x)$ & \textbf{0.825} & 0.489 & \textit{0.830*} & 0.731 & \textit{0.999*} & 0.811 & 0.653 \\
            $C(x)$ & 0.919 & 0.487 & 0.487 & \textit{0.969*} & \textit{0.999*} & 0.999 & \textbf{0.999} \\
            $\neg F(x,x)$ & \textbf{0.998} & \textit{0.998*} & 0.997 & 0.996 & 0.116 & \textit{0.989*} & 0.995 \\
            $F(x,y)\Rightarrow F(y,x)$ & 0.796 & \textit{0.852*} & 0.784 & 0.826 & 0.200 & \textit{0.778*} & \textbf{0.954} \\
            $\exists_y F(x,y)$ & 0.730 & \textit{0.764*} & 0.742 & \textbf{0.748} & 0.753 & \textit{0.715*} & 0.718 \\
            $F(x,y)\wedge S(x)\Rightarrow S(y)$ & \textbf{0.716} & 0.792 & \textit{0.656*} & 0.615 & 0.298 & \textit{0.742*} & 0.497 \\
            $S(x)\Rightarrow C(x)$ & 0.715 & 0.741 & 0.615 & \textit{0.806*} & 0.267 & 0.366 & \textit{\textbf{0.978}*} \\
            $\neg S(x)\Rightarrow\neg C(x)$ & 0.917 & 0.755 & 0.635 & \textit{0.910*} & 0.463 & 0.940 & \textit{\textbf{0.991}*} \\
            \midrule
            SAT of KB & \textbf{0.813} & 0.658 & 0.683 & 0.780 & 0.511 & 0.752 & 0.732 \\
            \bottomrule
        
        \end{tabular}
        \end{center}
    
    \end{table}


% The sources for the ceur-art style are available via
% \begin{itemize}
% \item \href{https://github.com/yamadharma/ceurart}{GitHub},
% % \item \href{https://www.overleaf.com/project/5e76702c4acae70001d3bc87}{Overleaf},
% \item
%   \href{https://www.overleaf.com/latex/templates/template-for-submissions-to-ceur-workshop-proceedings-ceur-ws-dot-org/pkfscdkgkhcq}{Overleaf
%     template}.
% \end{itemize}

\end{document}